\documentclass[11pt]{article}

\usepackage[final]{acl}

\usepackage{times}
\usepackage{latexsym}

\usepackage[T1]{fontenc}

\usepackage[utf8]{inputenc}

\usepackage{microtype}

\usepackage{inconsolata}

\usepackage{graphicx}
\usepackage{adjustbox}
\usepackage{subcaption}
\usepackage{booktabs}
\usepackage{multirow}
\usepackage{float}
\usepackage{pifont}
\usepackage{bbding}
\usepackage{latexsym}
\usepackage{wrapfig}
\usepackage{threeparttable}
\usepackage[skins, breakable]{tcolorbox}
\usepackage{booktabs,amsfonts,dcolumn}
\usepackage{tabularx}
\usepackage{colortbl}
\usepackage{algorithm} 
\usepackage{xcolor}
\usepackage{makecell}
\usepackage{amsmath}
\usepackage[utf8]{inputenc}
\usepackage{fvextra}
\newcommand{\headercolor}{\rowcolor{gray!15}}

\newtcolorbox{prbox}[1][]{
    enhanced,
    after skip=8mm, 
    title=#1,
    breakable = true,
    fonttitle=\sffamily\bfseries\color{white},
    coltitle=white,
    colbacktitle=gray!80!white, 
    titlerule=0pt,
    overlay={%
        \ifcase\tcbsegmentstate
        \or%
        \else%
        \fi%
    },
    colback = gray!2!white,          
    colframe = gray!80              
}

\title{Negative Advantage Is a Double-Edged Sword: \\Calibrating Advantages in GRPO for Search Agents}

\author{
Jiayi Wu\textsuperscript{1},
Ruobing Xie\textsuperscript{2,*}, 
Zeqian Huang\textsuperscript{3},
Lei Jiang\textsuperscript{3},  \\
\textbf{ 
Can Xu\textsuperscript{1},
Kangyang Luo\textsuperscript{4}, 
Bochen Lin\textsuperscript{1},
Ming Gao\textsuperscript{1},
Xiang Li\textsuperscript{1,*}}
\\
\textsuperscript{1}School of Data Science and Engineering, East China Normal University\\
\textsuperscript{2}Large Language Model Department, Tencent \quad
\textsuperscript{3}Yuanbao Team, Tencent\\
\textsuperscript{4}Shanghai University of Engineering Science \\
\tt{jiayiwu@stu.ecnu.edu.cn} \\
}

\begin{document}
\maketitle
\begingroup
\renewcommand{\thefootnote}{}
\footnotetext{\textsuperscript{*}Corresponding Authors}
\endgroup
\begin{abstract}
Search agents achieve strong question-answering performance through multi-turn interactions with search engines, with Group Relative Policy Optimization (GRPO) being a widely used training algorithm. However, GRPO-style algorithms still face several challenges in multi-hop search settings. First, correct intermediate steps are often penalized when the final answer is wrong. Second, training is highly unstable, often causing degradation of natural language ability or even catastrophic training collapse. Our analysis attributes these issues to coarse-grained advantage assignment and an imbalance between positive and negative advantages. To address these problems, we propose CalibAdv, an advantage calibration method specifically designed for search agents that enables more accurate and more stable modeling of penalties and rewards. Specifically, CalibAdv leverages the correctness of intermediate steps to downscale excessive negative advantages at a fine-grained level. It then further rebalances positive and negative advantages to improve training stability. Importantly, CalibAdv adopts a lightweight design that calibrates advantages from standard rollout signals, making it simple and easy to deploy. Extensive experiments across three models and seven benchmarks demonstrate that CalibAdv improves both model performance and training stability. Our code is available at \url{https://github.com/wujwyi/CalibAdv}.
\end{abstract}
\section{Introduction}

\begin{figure}[t]
    \centering
    \includegraphics[width=\linewidth]{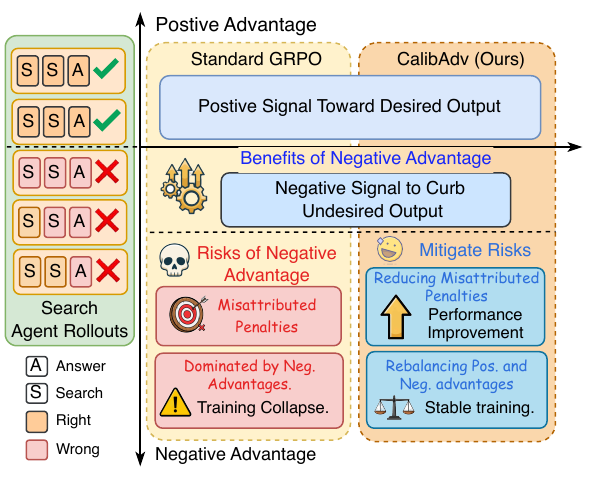}
    \caption{CalibAdv improves model performance and stabilizes training by scaling advantages.}
    \label{fig:intro}
\end{figure}

Search agents demonstrate impressive question-answering capabilities through iterative interactions with search engines \cite{shi2025deepresearchsystematicsurvey,Search-R1,zerosearch,ARR}. Compared with standard retrieval-augmented generation (RAG), search agents can flexibly reformulate search queries based on contextual information and adaptively determine the number of search iterations, enabling more effective resolution of multi-hop QA tasks. Their success critically relies on Reinforcement Learning with Verifiable Rewards (RLVR) such as GRPO \cite{DeepSeekMath}. 

Despite their widespread use in RLVR, GRPO-like methods remain challenging when training search agents.
\textbf{First, trajectory-level reward assignment can misalign with step-level correctness.} Search agents interact with search engines over multiple turns, but GRPO assigns advantages largely based on the final rollout outcome. Consequently, when the final answer is incorrect, correct or useful intermediate steps may still receive negative advantages, potentially hindering performance.
\textbf{Second, training instability is common} \cite{Search-R1,zerosearch,SimpleTIR, LLD}. As training proceeds, incoherent or nonsensical outputs become increasingly frequent, eventually leading to a catastrophic collapse of the model's natural language generation ability. This instability is more pronounced than in single-turn generation, as search agents repeatedly inject externally retrieved documents into the context across interaction turns~\cite{SimpleTIR}.

We observe that the \emph{\textbf{negative signals}}, as a key mechanism in reinforcement learning, are closely related to the issues described above. 
For intermediate-step mis-penalization, our analysis shows that this is not a rare corner case: \textbf{43.97\%} of penalized intermediate steps are close to correct, suggesting that advantages should be calibrated at a finer granularity according to step correctness. 
For training collapse, our analysis points to the persistent dominance of negative advantages: without sufficient positive advantages to anchor the policy, these negative updates can push it away from coherent outputs, thereby increasing entropy and perplexity and eventually causing collapse.
These observations inspire us to think: When training search agents with GRPO-like RL, although the penalty signal is crucial for success, \emph{negative advantages must be carefully calibrated based on the correctness of intermediate steps, and positive and negative advantages should be kept in balance.}

To this end, we propose \textbf{CalibAdv}, a GRPO advantage calibration method tailored for search agents that enables more accurate and more stable modeling of penalties and rewards. 
As illustrated in Figure~\ref{fig:intro}, CalibAdv adjusts advantages in three ways while following a simple first-principles design that uses signals already available in standard search agent rollouts. 
First, for intermediate steps, we introduce a simple and readily available rule-based silver document proxy to identify likely correct intermediate steps in incorrect rollouts and downscale their negative advantages.
Second, for the final answer step, which is particularly prone to imbalance, we rescale positive advantages to match the aggregate magnitude of negative advantages, thereby mitigating degradation in natural language generation and supporting stable long-term training without introducing an additional tunable scaling coefficient.
Finally, to eliminate the conflicting advantages induced by shared prefixes across positive and negative rollouts, we prepend such prefixes directly to the prompt and exclude them from advantage assignment, thereby mitigating format collapse.
Overall, CalibAdv remains simple and deployment friendly.

We conduct extensive experiments on three models and seven benchmarks, demonstrating that CalibAdv substantially improves both model performance and training stability. On average, CalibAdv yields an 11.80\% relative improvement in F1 score, while avoiding degradation of the model's natural language capabilities and training collapse. Our contributions can be summarized as follows:
\begin{itemize}
\item We mitigate the mismatch between intermediate step correctness and reward signals in GRPO-style training for multi-turn interactive tasks by introducing a simple and readily available rule-based silver document proxy for fine-grained advantage calibration.
\item We identify persistent negative-advantage dominance as a key factor associated with training collapse in GRPO-based search-agent training, and show that rebalancing positive and negative advantages effectively improves training stability with a natural 1.0 scaling, benefiting both future RL algorithm exploration and industrial-scale training.
\item Extensive experiments demonstrate the effectiveness and practicality of CalibAdv, highlighting that negative advantages are a double-edged sword: while they provide crucial supervisory signals, they can also cause performance degradation and training instability if left uncalibrated.
\end{itemize}

\section{Background}

\begin{figure*}[t]
    \centering
    \begin{subfigure}[b]{0.32\textwidth}
        \centering
        \includegraphics[width=\textwidth]{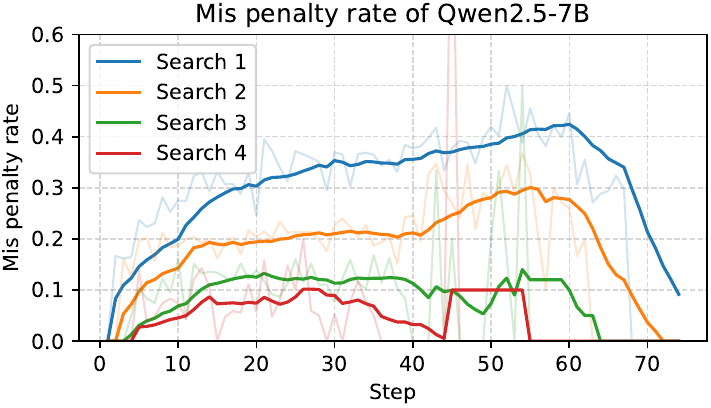}
    \end{subfigure}
    \begin{subfigure}[b]{0.32\textwidth}
        \centering
        \includegraphics[width=\textwidth]{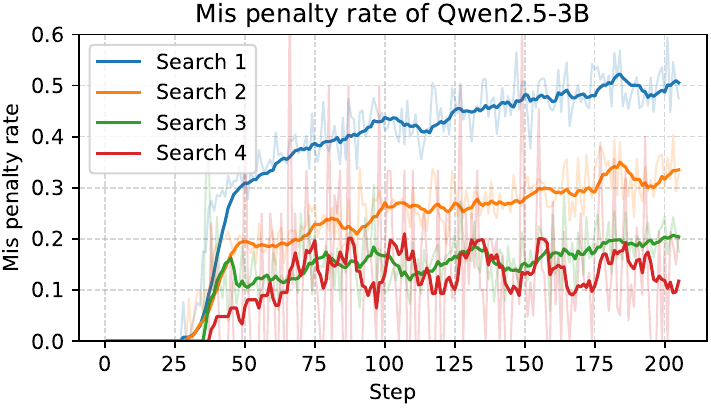}
    \end{subfigure}
    \begin{subfigure}[b]{0.32\textwidth}
        \centering
        \includegraphics[width=\textwidth]{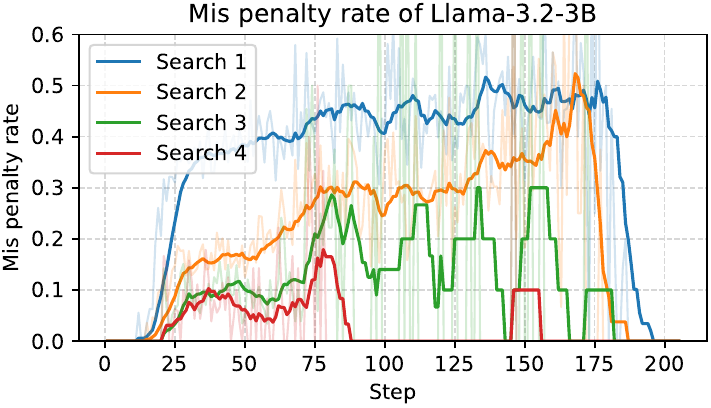}
    \end{subfigure}
    
    \caption{Proportion of erroneously penalized steps.}
    \label{fig:mis penalty}
\end{figure*}

In this section, we briefly introduce the pipeline of the multi-hop search task, as well as the reward design commonly adopted when training search agents with the GRPO algorithm.

\noindent
\textbf{The Multi-Hop Search Task Pipeline.}
\label{sec:multi-hop search pipeline}
A search agent answers questions via multi-turn interaction with a retriever \cite{Search-R1,R1-Searcher,zerosearch}. Given an input question or retrieved documents, the agent performs a reasoning process delimited by the \texttt{<think>} and \texttt{</think>} tags. After the reasoning process, if external information is required, the agent generates a retrieval query enclosed within the \texttt{<search>} and \texttt{</search>} tags. The retriever then returns relevant documents based on the generated query, wraps them with \texttt{<information>} and \texttt{</information>} tags, and appends them to the agent's input context. When sufficient information has been accumulated, the agent outputs the final answer following the reasoning process, enclosed within the \texttt{<answer>} and \texttt{</answer>} tags. Formally, given a question $x$, the agent $\pi$ generates the response at the $i$-th turn based on the context from the previous $i - 1$ turns, and then retrieves a document $d_i$ or produces the final answer:
\begin{equation}
y_i = \pi(x,y_1,d_1,y_2,d_2,\dots,y_{i-1},d_{i-1})
\end{equation}
The complete prompt is provided in Appendix~\ref{appendix:prompt}.

\noindent
\textbf{Reward Design for Search Agents.}
Following prior work \cite{R1-Searcher, zerosearch}, we adopt the word-level F1 score between the predicted answer and the ground truth as the answer reward $r_{answer}$, together with a binary format reward $r_{format} \in \{0, 1\}$ that indicates whether the output adheres to the required structure. The two rewards are combined via a gating mechanism, so that the answer reward is granted only when the format constraint is satisfied:
\begin{equation}
\begin{aligned}
r_{answer} &= \mathrm{F1}(\hat{y}, y), \\
r_{format} &=
\begin{cases}
1, & \text{format is correct,} \\
0, & \text{otherwise,}
\end{cases} \\
r_{final}  &= r_{answer} \cdot r_{format},
\end{aligned}
\end{equation}
where $\hat{y}$ and $y$ denote the predicted answer and the ground truth, respectively.

\section{Methodology}
\begin{figure*}[t]
    \centering
    \includegraphics[width=\textwidth]{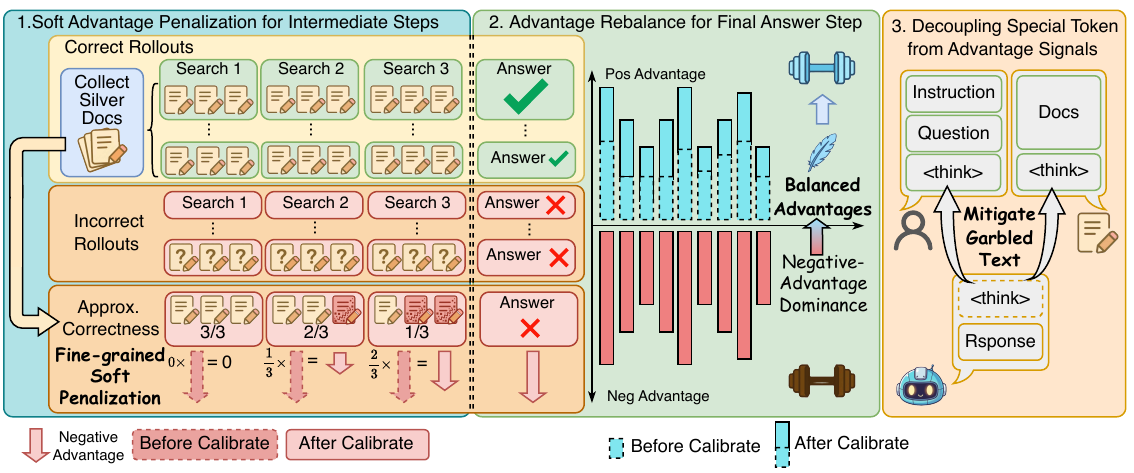}
    \caption{Overview of CalibAdv.}
    \label{fig:CalibAdv overview}
\end{figure*}

We begin with an empirical analysis of two key problems that arise when training search agents with standard GRPO, which directly motivate the design of CalibAdv: calibrating the advantage according to (i) the correctness of intermediate steps and (ii) the ratio of positive to negative advantages. An overview of CalibAdv is illustrated in Figure~\ref{fig:CalibAdv overview}.

\begin{figure*}[t]
    \centering
    \begin{subfigure}[b]{0.245\textwidth}
        \centering
        \includegraphics[width=\textwidth]{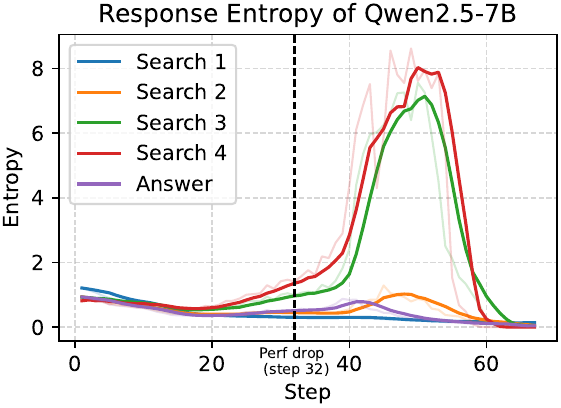}
    \end{subfigure}
    \begin{subfigure}[b]{0.245\textwidth}
        \centering
        \includegraphics[width=\textwidth]{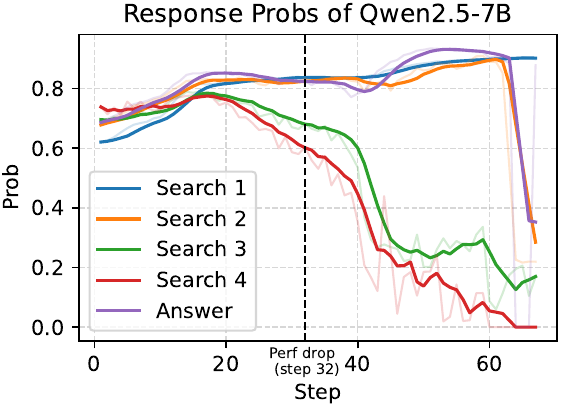}
    \end{subfigure}
    \begin{subfigure}[b]{0.245\textwidth}
        \centering
        \includegraphics[width=\textwidth]{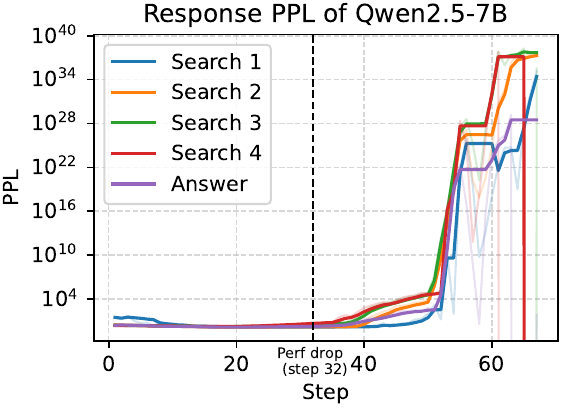}
    \end{subfigure}
    \begin{subfigure}[b]{0.245\textwidth}
        \centering
        \includegraphics[width=\textwidth]{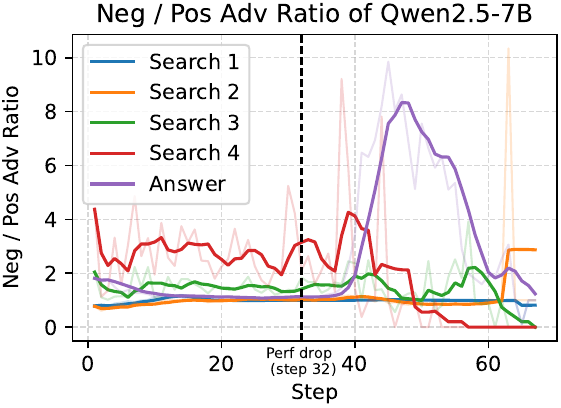}
    \end{subfigure}
    
    \caption{Training signals associated with language collapse.}
    \label{fig:collapse analysis}
\end{figure*}

\subsection{Existing Issues in Standard GRPO for Multi-Turn Search Agent Training}

\noindent
\textbf{Mis-Penalization of Correct Steps.}
\label{sec: mis penalization}
Since the search agent responds over multiple rounds (\S~\ref{sec:multi-hop search pipeline}), the trajectory space is large and the correctness of intermediate steps is non-uniform. However, standard GRPO propagates the advantage to all tokens within a rollout, assigning identical advantages to $y_1, y_2, \dots, y_n$. This causes a substantial mismatch between step-level correctness and the assigned advantages: if the final answer is wrong, many correct intermediate steps are erroneously penalized.

To verify this hypothesis, we examine how many intermediate steps assigned negative advantages are actually correct. As ground-truth correctness is infeasible to obtain, we adopt a proxy: an intermediate step assigned a negative advantage is regarded as \emph{likely correct} if all its retrieved documents are among the silver documents aggregated from correct rollouts. An empirical validation of the reliability of this proxy is provided in Appendix~\ref{appendix:silver_proxy_correctness}, and a detailed description of the implementation is provided in \S~\ref{sec: soft penalization}. As shown in Figure~\ref{fig:mis penalty}, experiments across multiple models reveal that a substantial portion of likely-correct steps are erroneously penalized. For example, on Qwen2.5-7B, \textbf{43.97\%} of penalized intermediate steps are likely correct and therefore erroneously penalized. Mis-penalization is more common in earlier rounds, whose contributions are more easily obscured by failures in subsequent steps. This indicates that the reward signal of standard GRPO in multi-turn search agent training suffers from significant latency and bias, making it crucial to calibrate negative advantages according to the correctness of intermediate reasoning steps.

\noindent
\textbf{Training Collapse.}
\label{sec: training collapse}
Prior work \cite{zerosearch,Search-R1,LLD} has observed that training search agents with GRPO is prone to training collapse. We find that beyond a certain point in training, the model's outputs first degenerate into mixtures of letters and Unicode symbols (``garbled text''), and eventually collapse into word-level repetition, dominated by \texttt{<think>}-related tokens or the Unicode replacement character (\texttt{\textbackslash ufffd}). Examples are shown in Appendix~\ref{appendix:garbled}.

To diagnose this language degradation, we monitor four training signals: output entropy, sampled-token probabilities, policy PPL, and the negative-to-positive advantage ratio. As shown in Figure~\ref{fig:collapse analysis}, capability decline is accompanied by an initial rise and subsequent collapse in entropy, lower token probabilities, sharply increased PPL, and persistently dominant negative advantages. In Appendix~\ref{appendix:round_dynamics}, we discuss the round-specific patterns and provide an empirically grounded explanation for why entropy declines after its rapid rise.

These observations indicate a strong connection between natural-language degradation and a persistent imbalance between positive and negative advantages. In policy-gradient optimization, negative advantages push the policy \emph{away from} sampled outputs, whereas positive advantages anchor it \emph{toward} desirable regions \cite{zhu2025negative}. 
As shown in Figure~\ref{fig:collapse analysis}, negative advantages dominate over positive ones throughout almost the entire training process. Rollouts that exhibit well-formed natural language structure but yield incorrect final answers are repeatedly pushed away from, while the policy lacks sufficient positive signals to anchor it within the region of coherent language.
We hypothesize that this persistent imbalance gradually flattens the policy distribution, as reflected by rising entropy and PPL, and contributes to the eventual collapse.
This motivates calibrating the advantage to prevent such imbalance between positive and negative advantages throughout training, which we identify as a critical step toward stable optimization.

\subsection{Soft Advantage Penalization for Intermediate Steps}
\label{sec: soft penalization}
To mitigate the mis-penalization identified above, we calibrate negative advantages with a step-wise correctness signal. The key challenge lies in selecting such a signal that is both readily accessible and sufficiently accurate.

In multi-hop search, the most essential function of each intermediate step is to reasonably integrate the preceding context and formulate an appropriate query, so as to retrieve informative documents that could help derive the final answer. Building on this insight, for each question, the documents retrieved by correct rollouts are considered as silver documents. If a penalized intermediate step retrieves any silver documents for the corresponding question, it is regarded as having made a meaningful advance toward the final answer with high probability, thereby containing likely-correct information. Intuitively, the correctness score of an intermediate step is related to the proportion of retrieved documents that are silver. This approach leverages the intrinsic properties of the multi-hop search task, avoiding reliance on external LLMs or additional sampling. Formally, let $\mathcal{R}_q^{\text{correct}}$ denote the set of correct rollouts for question $q$, and let $\mathcal{D}_r$ denote the set of documents retrieved by rollout $r$. The operational criterion for selecting correct rollouts is provided in Appendix~\ref{appendix:soft_penalty_impl}. The set of silver documents is defined as follows:
\begin{equation}
\mathcal{D}_q^{\text{silver}} = \bigcup_{r \in \mathcal{R}_q^{\text{correct}}} \mathcal{D}_r.
\end{equation}
Let $\mathcal{D}_s$ denote the set of documents retrieved by an intermediate step $s$. The correctness score $c_s$ is then computed as follows:
\begin{equation}
c_s = \frac{|\mathcal{D}_s \cap \mathcal{D}_q^{\text{silver}}|}{|\mathcal{D}_s|}, \quad c_s \in [0,1].
\label{eq:cs_score}
\end{equation}
To prevent the model from exploiting soft penalization by repeatedly issuing similar queries, we introduce a repetition-aware gating mechanism that suppresses soft-penalty correction for repeated or near-duplicate queries before advantage calibration. The detailed gating criterion is provided in Appendix~\ref{appendix:soft_penalty_impl}.

Once the correctness of intermediate steps is obtained, we attenuate the magnitude of erroneously assigned negative advantages to constrain erroneous penalties. Unlike standard GRPO, which assigns a single advantage uniformly to every token within a rollout, we treat each intermediate step as the basic unit for advantage adjustment. Concretely, an intermediate step $s$ corresponds to a single turn $y_i$ in the multi-hop search pipeline (\S~\ref{sec:multi-hop search pipeline}), comprising the reasoning segment within \texttt{<think>}\,\dots\,\texttt{</think>} and the retrieval query within \texttt{<search>}\,\dots\,\texttt{</search>} produced at that turn (excluding the final-answer turn). The adjustment is performed independently for each intermediate step according to its own correctness score $c_s$, and the resulting $\tilde{A}_s$ is broadcast to every token within that step. Formally, let $A_s$ denote the original advantage of an intermediate step $s$. The adjusted advantage $\tilde{A}_s$ is defined as:
\begin{equation}
\tilde{A}_s =
\begin{cases}
A_s \cdot (1 - c_s), & \text{if } A_s < 0, \\
A_s, & \text{otherwise}.
\end{cases}
\end{equation}
In this way, the negative advantage of an intermediate step is attenuated in proportion to its likely correctness. The attenuation factor is directly determined by $1-c_s$, without introducing an additional penalty coefficient.

\subsection{Advantage Rebalancing for the Final Answer Step}
\label{sec: rebalance}
To address the long-term dominance of negative advantages that degrades the model's natural language ability, we propose an advantage rebalancing mechanism for the final answer step, which contributes most to such dominance, so as to restore effective guidance from positive advantages. For intermediate steps, the soft penalization introduced above already allows positive advantages to play the dominant guiding role.

Concretely, for each group we rescale the positive advantages at the final answer step by the ratio between the magnitudes of the negative and positive advantages, so as to recalibrate their relative influence on the policy update. Formally, let $A_g^+$ and $A_g^-$ denote the positive and negative advantages at the final answer step of group $g$, respectively. We define the group-wise rebalancing ratio as
\begin{equation}
r_g = |A_g^-|/A_g^+,
\end{equation}
and obtain the rebalanced positive advantage by
\begin{equation}
\tilde{A}_g^+ = r_g \cdot A_g^+.
\end{equation}
By construction, this rescaling matches the aggregate magnitudes of positive and negative advantages and therefore introduces no additional scaling coefficient. We analyze sensitivity to deviations from this exact-balancing rule in \S~\ref{sec:lambda}. This rebalancing mitigates the dominance of negative advantages and stabilizes optimization.

\subsection{Decoupling the Special Token from Advantage Signals}
\label{sec: prepend think token}

As discussed in Section~\ref{sec: training collapse}, the word-level repetition that emerges after the collapse of the model's natural language ability is often associated with the \texttt{<think>} token. Moreover, we observe that the tokens most affected by the GRPO clipping mechanism are mainly related to \texttt{<think>} (see Appendix~\ref{appendix:clipped_tokens} for detailed statistics). Since \texttt{<think>} serves as a shared prefix of both positive and negative rollout trajectories, it receives massive conflicting training signals despite its fixed position, leading to large probability fluctuations. To prevent such fluctuations from triggering format collapse, we explicitly prepend \texttt{<think>} after the prompt and external documents, thereby excluding it from the tokens optimized by the policy.

\section{Experiments}
\begin{table*}[t]
\centering
\resizebox{1.0\linewidth}{!}{
\begin{tabular}{lccccccccc}
\toprule
\textbf{Datasets} &
\multicolumn{3}{c}{\textbf{Single-Hop QA}}  & 
\multicolumn{4}{c}{\textbf{Multi-Hop QA}} & 
\multirow{2}{*}{\textbf{Average}} &
\multirow{2}{*}{\textbf{Collapse Point}}\\
 \cmidrule(lr){2-4}  \cmidrule(lr){5-8} 

{\textbf{Models}} & {\textbf{NQ}} & \textbf{TriviaQA} & \textbf{PopQA} & {\textbf{HotpotQA}} & {\textbf{2Wiki}}  
& {\textbf{MuSiQue}} & {\textbf{Bamboogle}} \\

\midrule
\headercolor
\multicolumn{10}{c}{\textbf{\textit{Qwen2.5-7B}}} \\
Search-R1 & 45.70 & 70.78 & 50.09 & 50.16 & 53.00 & 25.65 & 48.66 & 49.15 & 58/205 \\
SimpleTIR & 48.13 & 69.34 & 47.84 & 54.47 & 49.95 & 26.25 & 50.90 & 49.55 & 65/205 \\
LLD & 52.23 & 74.24 & 49.70 & 60.63 & 56.17 & 31.57 & 58.36 & 54.70 & No collapse \\
StepSearch & 45.76 & 66.40 & 45.86 & 48.84 & 46.94 & 29.01 & 55.74 & 48.36 & --\textsuperscript{$\ddagger$} \\
MT-GRPO & 51.99 & 75.57 & 50.86 & 54.00 & 46.46 & 23.79 & 51.47 & 50.59 & No collapse\textsuperscript{$\dagger$} \\
Tree-GRPO & 51.78 & 72.96 & 48.95 & 57.67 & 51.16 & 29.59 & 55.21 & 52.47 & 147/205\\
GiGPO & 47.19 & 66.52 & 46.50 & 50.64 & 47.79 & 28.50 & 47.11 & 47.79 & No collapse\textsuperscript{$\dagger$} \\
CalibAdv (Ours) & \textbf{53.20} & \textbf{75.71} & \textbf{52.40} & \textbf{63.09} & \textbf{56.91} & \textbf{33.36} & \textbf{62.25} & \textbf{56.70} & \textbf{No collapse} \\
\midrule
\headercolor
\multicolumn{10}{c}{\textbf{\textit{Qwen2.5-3B}}\textsuperscript{$\S$}} \\
Search-R1 & 47.83 & 69.58 & 44.88 & 56.52 & 49.83 & 28.33 & 50.48 & 49.64 & 292/410 \\
SimpleTIR & 47.31 & 70.39 & 44.74 & 56.54 & 48.75 & 25.83 & 48.58 & 48.88 & 310/410 \\
CalibAdv (Ours) & \textbf{50.95} & \textbf{72.03} & \textbf{50.48} & \textbf{59.19} & \textbf{52.51} & \textbf{30.07} & \textbf{50.62} & \textbf{52.26} & \textbf{No collapse} \\
\midrule
\headercolor
\multicolumn{10}{c}{\textbf{\textit{Llama-3.2-3B-Instruct}}} \\
Search-R1 & 44.74 & 68.56 & 44.45 & 48.08 & 40.19 & 19.92 & 47.33 & 44.75 & 40/205 \\
SimpleTIR & 47.39 & 69.64 & 46.13 & 49.60 & 42.84 & 21.85 & 45.93 & 46.20 & 54/205 \\
LLD & 47.98 & 70.32 & 47.88 & 55.59 & \textbf{47.80} & 23.70 & 53.34 & 49.52 & No collapse \\
CalibAdv (Ours) & \textbf{50.75} & \textbf{72.37} & \textbf{50.98} & \textbf{56.97} & 47.36 & \textbf{26.77} & \textbf{54.20} & \textbf{51.34} & \textbf{No collapse} \\

\bottomrule
\end{tabular}
}
\begin{flushleft}
\footnotesize
$\dagger$ MT-GRPO performs a single retrieval and GiGPO averages 1.28 searches, so neither exhibits multi-step search collapse.
$\ddagger$ StepSearch is trained with PPO rather than GRPO, and thus does not exhibit GRPO-specific collapse.
$\S$ LLD did not produce a usable checkpoint on Qwen2.5-3B and is omitted.
\end{flushleft}
\caption{Main results on F1 score and Collapse Point. The best result for each model is shown in \textbf{bold}.}
\label{tab:main res}
\end{table*}

\subsection{Experimental Setup}
\subsubsection{Dataset and Evaluation Methodology}

\paragraph{Datasets}
We construct the training set from the training splits of two multi-hop QA datasets, HotpotQA \cite{HotpotQA} and 2WikiMultiHopQA \cite{2wiki}.
For evaluation, we use the corresponding test splits and additionally include Natural Questions \cite{nq}, TriviaQA \cite{TriviaQA}, PopQA \cite{popqa}, MuSiQue \cite{MuSiQue}, and Bamboogle \cite{bam} to assess out-of-domain generalization. Natural Questions, TriviaQA, and PopQA are single-hop benchmarks, whereas MuSiQue and Bamboogle are multi-hop benchmarks. Detailed dataset statistics are provided in Appendix~\ref{appendix:data_stats}.

\paragraph{Evaluation Metrics}
We evaluate both answer accuracy and training stability, reflecting our goals of improving performance while mitigating training collapse. Accuracy is measured by the \textbf{F1 score} used in the reward design. Stability is measured by the \textbf{Collapse Point}, where a later point indicates more stable training. Its formal definition and the corresponding checkpoint-selection rule are provided in Appendix~\ref{appendix:collapse_definition}.

\subsubsection{Implementation Details}
We use three widely used large language models as the backbone models for search agents: Qwen2.5-3B \cite{Qwen2.5}, Qwen2.5-7B, and Llama-3.2-3B-Instruct. Following \citet{Search-R1}, we use the Wikipedia dump from December 20, 2018 \cite{wiki2018} as the retrieval corpus, with E5-Base-V2 \cite{E5} as the dense retriever and ms-marco-MiniLM-L12-v2\footnote{\url{https://huggingface.co/cross-encoder/ms-marco-MiniLM-L12-v2}} as the reranker. For each search query, the retriever returns the top-20 documents, which are reranked to select the top-3. We perform full fine-tuning of the search agent, with training hyperparameters detailed in Appendix~\ref{appendix:Hyperparameters}.

\subsubsection{Baselines}
We compare CalibAdv with three categories of baselines. The standard GRPO baseline is \textbf{Search-R1} \cite{Search-R1}. Collapse-mitigation baselines include \textbf{SimpleTIR} \cite{SimpleTIR} and \textbf{LLD} \cite{LLD}. Baselines that introduce process-level training signals include \textbf{StepSearch} \cite{StepSearch}, \textbf{MT-GRPO} \cite{MTGRPO}, \textbf{Tree-GRPO} \cite{TreeGRPO}, and \textbf{GiGPO} \cite{GiGPO}. Detailed implementations are provided in Appendix~\ref{appendix:baseline_impl}. To ensure a fair comparison, all baselines use the same training data and retrieval pipeline as CalibAdv.

\begin{figure*}[t]
    \centering
    \begin{subfigure}[b]{0.32\textwidth}
        \centering
        \includegraphics[width=\textwidth]{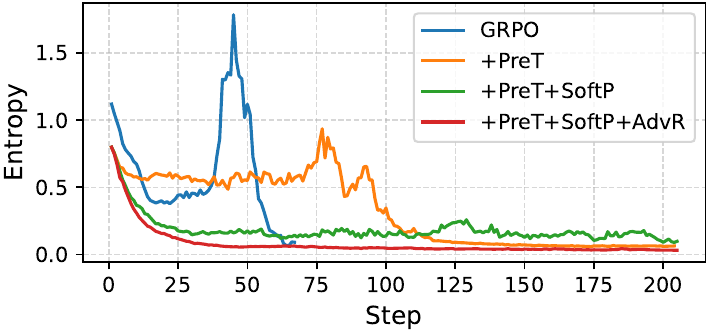}
    \end{subfigure}
    \hfill
    \begin{subfigure}[b]{0.32\textwidth}
        \centering
        \includegraphics[width=\textwidth]{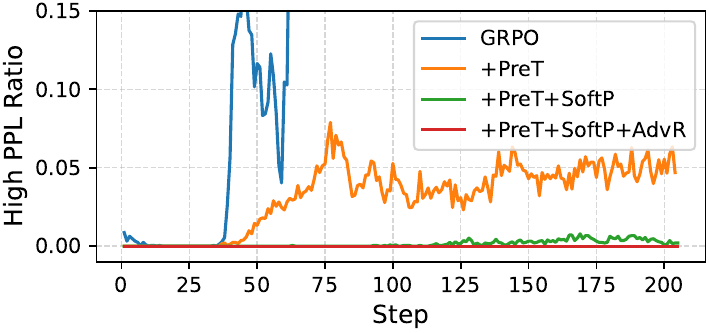}
    \end{subfigure}
    \hfill
    \begin{subfigure}[b]{0.32\textwidth}
        \centering
        \includegraphics[width=\textwidth]{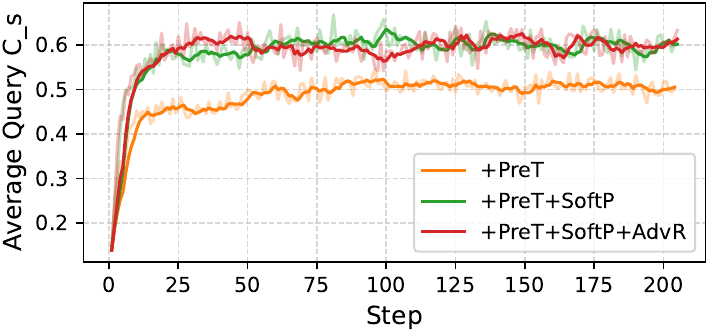}
    \end{subfigure}

    \begin{subfigure}[b]{0.32\textwidth}
        \centering
        \includegraphics[width=\textwidth]{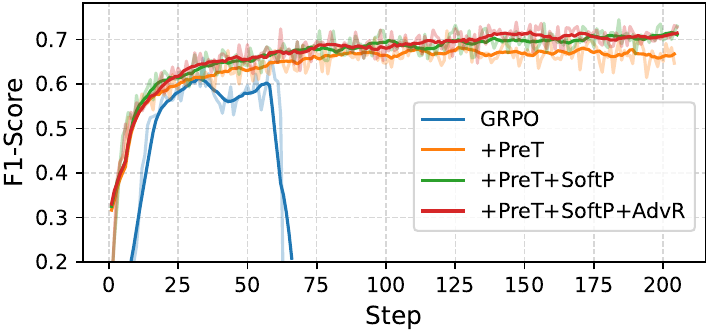}
    \end{subfigure}
    \hfill
    \begin{subfigure}[b]{0.32\textwidth}
        \centering
        \includegraphics[width=\textwidth]{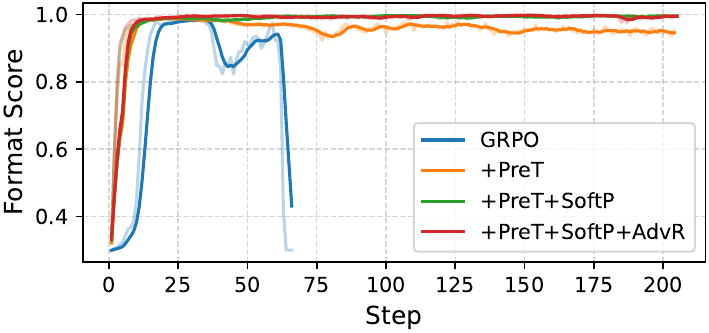}
    \end{subfigure}
    \hfill
    \begin{subfigure}[b]{0.32\textwidth}
        \centering
        \includegraphics[width=\textwidth]{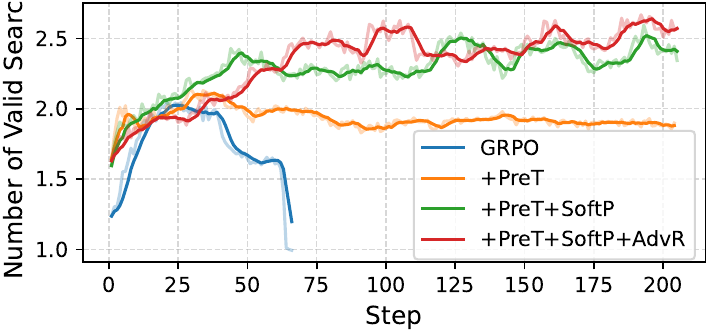}
    \end{subfigure}

    \caption{Training dynamics as the components of CalibAdv are progressively introduced.}
    \label{fig:ablation curve}
\end{figure*}
\begin{table*}[t]
\centering
\resizebox{1.0\linewidth}{!}{
\begin{tabular}{lcccccccc}
\toprule

{\textbf{Models}} & {\textbf{NQ}} & \textbf{TriviaQA} & \textbf{PopQA} & {\textbf{HotpotQA}} & {\textbf{2Wiki}}  
& {\textbf{MuSiQue}} & {\textbf{Bamboogle}} & \textbf{Average}\\

\midrule
\headercolor
\multicolumn{9}{c}{\textbf{\textit{F1 Score}}($\uparrow$)} \\
Standard GRPO & 45.70 & 70.78 & 50.09 & 50.16 & 53.00 & 25.65 & 48.66 & 49.15 \\
+PreThink & 49.87 & 70.89 & 47.73 & 59.37 & 52.66 & 30.37 & 57.55 & 52.63 \\
+PreThink+SoftPen  & 52.93 & 75.05 & 51.41 & \textbf{63.13} & 56.64 & \textbf{33.90} & 60.70 & 56.25 \\
+PreThink+SoftPen+AdvReb & \textbf{53.20} & \textbf{75.71} & \textbf{52.40} & 63.09 & \textbf{56.91} & 33.36 & \textbf{62.25} & \textbf{56.70} \\
\midrule
\headercolor
\multicolumn{9}{c}{\textbf{\textit{High PPL Ratio}}($\downarrow$)} \\
Standard GRPO & \multicolumn{8}{c}{Training Collapse}\\
+PreThink & 8.39 & 7.38 & 9.94 & 6.06 & 6.80 & 10.59 & 2.42 & 7.37 \\
+PreThink+SoftPen & 7.97 & 6.61 & 7.08 & 3.85 & 4.93 & 10.55 & 5.65 & 6.66 \\
+PreThink+SoftPen+AdvReb & \textbf{0.00} & \textbf{0.00} & \textbf{0.00} & \textbf{0.00} & \textbf{0.00} & \textbf{0.00} & \textbf{0.00} & \textbf{0.00} \\

\bottomrule
\end{tabular}
}
\caption{Ablation study of CalibAdv on Qwen2.5-7B. PreThink denotes prepending the \texttt{<think>} token, SoftPen denotes soft penalization for intermediate steps, and AdvReb denotes advantage rebalancing at the final answer step. The best results are marked in \textbf{bold}.}
\label{tab:ablation f1}
\end{table*}

\subsection{Main Results}
\label{sec:main res}
The experimental results of CalibAdv and the baselines on seven benchmarks are presented in Table~\ref{tab:main res}. CalibAdv consistently outperforms all baselines and maintains stable training without any significant performance drop. Compared to standard GRPO (i.e., Search-R1), CalibAdv achieves an average relative improvement of 11.80\% in F1 score. 

CalibAdv also achieves better performance compared with baselines that rely on process-level training signals, while operating under a different design choice: it does not introduce the additional intermediate annotations used by StepSearch, nor the complex tree/group construction used by MT-GRPO, Tree-GRPO, and GiGPO. Instead, the silver-document proxy provides a reliable and simple signal for soft penalization, without extra annotation cost or complex sampling overhead.

Among baselines designed to stabilize training, SimpleTIR delays collapse but does not address the dominance of negative advantages. LLD avoids collapse on Qwen2.5-7B and Llama-3.2-3B-Instruct, yet still underperforms CalibAdv on both models. On the weaker Qwen2.5-3B, LLD fails to enter effective training altogether. We attribute this to its regularization term, which overly restricts the model's exploration and traps it in a local optimum. CalibAdv instead calibrates the advantage signal rather than relying on post-hoc constraints as in LLD or SimpleTIR, and remains robust across models of varying capability.

\subsection{Ablation Study}

We conduct ablation studies to evaluate the effectiveness of each component in CalibAdv. To quantify the severity of natural language degradation, we introduce the \textbf{High PPL Ratio}, defined as the proportion of outputs whose PPL $>$ 50\footnote{Empirically, outputs with a PPL $>$ 50 are almost nonsensical, typically exhibiting garbled text or word-level repetition.}. A higher value indicates more severe degradation in natural language generation. Table~\ref{tab:ablation f1} reports the ablation results, and Figure~\ref{fig:ablation curve} shows the training dynamics after progressively incorporating each component of CalibAdv, including entropy, average query correctness score, high PPL ratio, F1 score, format score, and number of valid search steps. 

Overall, each stage improves performance and stability with different emphases. Specifically, prepending \texttt{<think>} improves stability and enables a full epoch, but entropy still rises and 7.3\% of outputs remain high-PPL, showing that mandatory shared prefixes in GRPO rollouts can induce generation degradation. Soft penalization slightly lowers the High PPL Ratio and substantially raises the average query correctness score, yielding markedly better F1 and more valid search steps by avoiding over-penalization of correct intermediate reasoning. Finally, final-step advantage rebalancing eliminates high-PPL outputs and further improves F1, completely resolving natural language degradation.

\subsection{Impact of the Rebalancing Scaling Coefficient $\lambda$}
\label{sec:lambda}
\begin{figure}[t]
    \centering
    \begin{subfigure}[b]{0.45\textwidth}
        \centering
        \includegraphics[width=\textwidth]{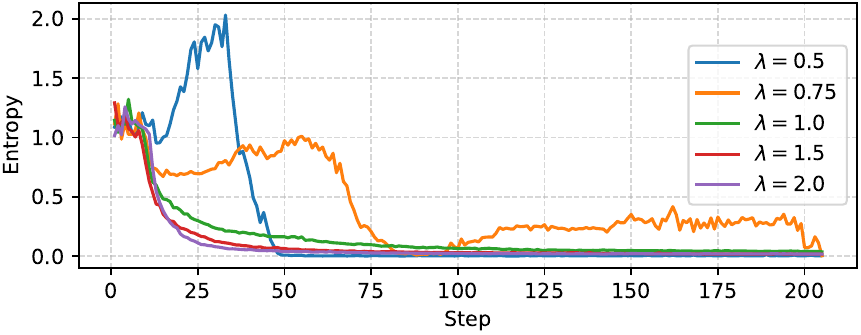}
    \end{subfigure}
    \begin{subfigure}[b]{0.45\textwidth}
        \centering
        \includegraphics[width=\textwidth]{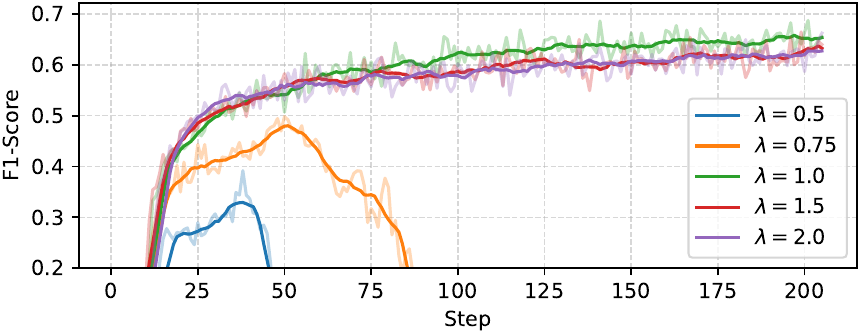}
    \end{subfigure}
    \vspace{-1pt}
    \caption{Impact of the rebalancing scaling coefficient $\lambda$ on performance and entropy dynamics.}
    \label{fig:lambda}
\end{figure}

To examine sensitivity to deviations from exact balancing, we generalize the default rule as $\tilde{A}_g^+=\lambda r_g A_g^+$, where $\lambda=1.0$ recovers the default CalibAdv formulation. We conduct this analysis on Qwen2.5-3B. To isolate the effect of $\lambda$, we remove the prepended think token and soft advantage penalization, and apply rebalancing at the same step within each group. As shown in Figure~\ref{fig:lambda}, $\lambda=1.0$ achieves the best performance, while either increasing or decreasing it degrades results. Larger $\lambda$ values make training dominated by positive advantages, which stabilizes learning but suppresses exploration and lowers the reward ceiling. Smaller values amplify the effect of negative advantages, causing language capability degradation and eventual collapse. These results support exact balancing as the default formulation and highlight the importance of balancing positive and negative advantages for stable and effective training.
\vspace{-4pt}
\subsection{Reliability of the Silver Document Proxy}
We further validate whether $c_s$ reliably tracks the helpfulness of intermediate steps. Using DeepSeek-V3.2 to evaluate 100 sampled steps at each score level, the helpful ratio increases monotonically from 17\% at $c_s=0$ to 39\%, 59\%, and 83\% at $c_s=1/3$, $2/3$, and $1$, respectively. For steps with $c_s=1$, a separate human evaluation yields 89\% helpfulness. Moreover, replacing our proxy with DeepSeek-V3.2-based step scoring yields negligible performance difference while incurring substantially higher training cost. Detailed validation protocols and results are provided in Appendix~\ref{appendix:silver_proxy_correctness}.

\section{Related Work}
\subsection{Training Collapse in GRPO-Like Algorithms for Multi-Turn Tasks}

Many GRPO-style methods for training search agents suffer from training collapse \cite{Search-R1,zerosearch}, and recent studies have proposed various mitigation strategies. Some studies analyze numerical instability as a key factor \cite{DefeatingFP16}, while others attribute collapse to out-of-distribution rollouts \cite{SimpleTIR} or sparse multi-turn rewards \cite{RAGEN} and address it through rollout or reward-group filtering. These methods are not specifically designed for multi-hop search and thus do not directly target its characteristics. More recently, LLD \cite{LLD} focuses on the multi-hop search setting and introduces an additional regularization term that prevents the likelihood of non-negative-advantage responses from being decreased. However, such a constraint may limit exploration during training.

\subsection{Intermediate Supervision for Multi-Turn GRPO}
Sparse reward signals for intermediate reasoning steps in multi-hop problems remain a long-standing challenge. Annotation-based methods, such as StepSearch \cite{StepSearch} and CriticSearch \cite{CriticSearch}, leverage gold intermediate answers or external critics to provide step-level signals, typically relying on dataset-specific annotations or auxiliary supervision models. Tree/group-based methods, including MT-GRPO \cite{MTGRPO}, Tree-GRPO \cite{TreeGRPO}, and GiGPO \cite{GiGPO}, estimate process-level advantages from intermediate comparison groups built upon tree rollouts or repeated states, which involves relatively complex sampling and group construction.
In contrast, CalibAdv leverages the silver-document proxy as a reliable and simple signal for step-level advantage calibration, without extra intermediate annotations or tree/group construction.
\section{Conclusion}
In this paper, we propose CalibAdv, a simple advantage calibration framework for GRPO training of search agents. CalibAdv combines soft penalization with silver documents, advantage rebalancing at the final answer step, and special-token decoupling, using only readily available rollout artifacts and requiring no process annotations, external LLM supervision, or tree/group construction. These properties make CalibAdv practical to deploy in large-scale search-agent training. Experiments on three models and seven benchmarks show that CalibAdv improves question-answering performance while maintaining stable training.

\section*{Limitations}

Due to computational resource constraints, we evaluate CalibAdv only in the Wikipedia-based search environment and leave broader web-search settings for future work.

\section*{Ethics Statement}
This work was conducted in strict compliance with the ACL Ethics Policy.
All datasets and models used in the experiments are publicly available.
The human evaluation was conducted internally by the authors on 100 samples; no external annotators were recruited, no personal data were collected, and no payment was involved.
Furthermore, our work aims to explore an advantage calibration method for training search agents with GRPO-style reinforcement learning.
We do not foresee any negative ethical impacts arising from our work.

\section*{Acknowledgments}

This work was supported by the Young Elite Scientists Sponsorship Program by CAST (2023QNRC001).

\bibliography{custom}

@article{zerosearch,
  author       = {Hao Sun and
                  Zile Qiao and
                  Jiayan Guo and
                  Xuanbo Fan and
                  Yingyan Hou and
                  Yong Jiang and
                  Pengjun Xie and
                  Yan Zhang and
                  Fei Huang and
                  Jingren Zhou},
  title        = {{ZeroSearch}: Incentivize the Search Capability of {LLMs} without Searching},
  journal      = {CoRR},
  volume       = {abs/2505.04588},
  year         = {2025},
  url          = {https://doi.org/10.48550/arXiv.2505.04588},
  doi          = {10.48550/ARXIV.2505.04588},
  eprinttype    = {arXiv},
  eprint       = {2505.04588},
  bibsource    = {dblp computer science bibliography, https://dblp.org}
}

@article{R1-Searcher,
  author       = {Huatong Song and
                  Jinhao Jiang and
                  Yingqian Min and
                  Jie Chen and
                  Zhipeng Chen and
                  Wayne Xin Zhao and
                  Lei Fang and
                  Ji{-}Rong Wen},
  title        = {{R1-Searcher}: Incentivizing the Search Capability in {LLMs} via Reinforcement
                  Learning},
  journal      = {CoRR},
  volume       = {abs/2503.05592},
  year         = {2025},
  url          = {https://doi.org/10.48550/arXiv.2503.05592},
  doi          = {10.48550/ARXIV.2503.05592},
  eprinttype    = {arXiv},
  eprint       = {2503.05592},
  bibsource    = {dblp computer science bibliography, https://dblp.org}
}

@misc{Search-R1,
      title={{Search-R1}: Training {LLMs} to Reason and Leverage Search Engines with Reinforcement Learning}, 
      author={Bowen Jin and Hansi Zeng and Zhenrui Yue and Jinsung Yoon and Sercan Arik and Dong Wang and Hamed Zamani and Jiawei Han},
      year={2025},
      eprint={2503.09516},
      archivePrefix={arXiv},
      primaryClass={cs.CL},
      url={https://arxiv.org/abs/2503.09516}, 
}

@misc{DefeatingFP16,
      title={Defeating the Training-Inference Mismatch via {FP16}}, 
      author={Penghui Qi and Zichen Liu and Xiangxin Zhou and Tianyu Pang and Chao Du and Wee Sun Lee and Min Lin},
      year={2025},
      eprint={2510.26788},
      archivePrefix={arXiv},
      primaryClass={cs.LG},
      url={https://arxiv.org/abs/2510.26788}, 
}

@misc{SimpleTIR,
      title={{SimpleTIR}: End-to-End Reinforcement Learning for Multi-Turn Tool-Integrated Reasoning}, 
      author={Zhenghai Xue and Longtao Zheng and Qian Liu and Yingru Li and Xiaosen Zheng and Zejun Ma and Bo An},
      year={2025},
      eprint={2509.02479},
      archivePrefix={arXiv},
      primaryClass={cs.LG},
      url={https://arxiv.org/abs/2509.02479}, 
}

@misc{RAGEN,
      title={{RAGEN}: Understanding Self-Evolution in {LLM} Agents via Multi-Turn Reinforcement Learning}, 
      author={Zihan Wang and Kangrui Wang and Qineng Wang and Pingyue Zhang and Linjie Li and Zhengyuan Yang and Xing Jin and Kefan Yu and Minh Nhat Nguyen and Licheng Liu and Eli Gottlieb and Yiping Lu and Kyunghyun Cho and Jiajun Wu and Li Fei-Fei and Lijuan Wang and Yejin Choi and Manling Li},
      year={2025},
      eprint={2504.20073},
      archivePrefix={arXiv},
      primaryClass={cs.LG},
      url={https://arxiv.org/abs/2504.20073}, 
}

@misc{LLD,
      title={On {GRPO} Collapse in {Search-R1}: The Lazy Likelihood-Displacement Death Spiral}, 
      author={Wenlong Deng and Yushu Li and Boying Gong and Yi Ren and Christos Thrampoulidis and Xiaoxiao Li},
      year={2025},
      eprint={2512.04220},
      archivePrefix={arXiv},
      primaryClass={cs.CL},
      url={https://arxiv.org/abs/2512.04220}, 
}

@misc{shi2025deepresearchsystematicsurvey,
      title={Deep Research: A Systematic Survey}, 
      author={Zhengliang Shi and Yiqun Chen and Haitao Li and Weiwei Sun and Shiyu Ni and Yougang Lyu and Run-Ze Fan and Bowen Jin and Yixuan Weng and Minjun Zhu and Qiujie Xie and Xinyu Guo and Qu Yang and Jiayi Wu and Jujia Zhao and Xiaqiang Tang and Xinbei Ma and Cunxiang Wang and Jiaxin Mao and Qingyao Ai and Jen-Tse Huang and Wenxuan Wang and Yue Zhang and Yiming Yang and Zhaopeng Tu and Zhaochun Ren},
      year={2025},
      eprint={2512.02038},
      archivePrefix={arXiv},
      primaryClass={cs.CL},
      url={https://arxiv.org/abs/2512.02038}, 
}

@misc{DeepSeekMath,
      title={{DeepSeekMath}: Pushing the Limits of Mathematical Reasoning in Open Language Models}, 
      author={Zhihong Shao and Peiyi Wang and Qihao Zhu and Runxin Xu and Junxiao Song and Xiao Bi and Haowei Zhang and Mingchuan Zhang and Y. K. Li and Y. Wu and Daya Guo},
      year={2024},
      eprint={2402.03300},
      archivePrefix={arXiv},
      primaryClass={cs.CL},
      url={https://arxiv.org/abs/2402.03300}, 
}

@misc{StepSearch,
      title={{StepSearch}: Igniting {LLMs}' Search Ability via Step-Wise Proximal Policy Optimization}, 
      author={Ziliang Wang and Xuhui Zheng and Kang An and Cijun Ouyang and Jialu Cai and Yuhang Wang and Yichao Wu},
      year={2025},
      eprint={2505.15107},
      archivePrefix={arXiv},
      primaryClass={cs.CL},
      url={https://arxiv.org/abs/2505.15107}, 
}

@misc{MTGRPO,
      title={Reinforcing Multi-Turn Reasoning in {LLM} Agents via Turn-Level Reward Design}, 
      author={Quan Wei and Siliang Zeng and Chenliang Li and William Brown and Oana Frunza and Wei Deng and Anderson Schneider and Yuriy Nevmyvaka and Yang Katie Zhao and Alfredo Garcia and Mingyi Hong},
      year={2025},
      eprint={2505.11821},
      archivePrefix={arXiv},
      primaryClass={cs.LG},
      url={https://arxiv.org/abs/2505.11821}, 
}

@misc{CriticSearch,
      title={{CriticSearch}: Fine-Grained Credit Assignment for Search Agents via a Retrospective Critic}, 
      author={Yaocheng Zhang and Haohuan Huang and Zijun Song and Yuanheng Zhu and Qichao Zhang and Zijie Zhao and Dongbin Zhao},
      year={2025},
      eprint={2511.12159},
      archivePrefix={arXiv},
      primaryClass={cs.CL},
      url={https://arxiv.org/abs/2511.12159}, 
}

@misc{Qwen2.5,
      title={Qwen2.5 Technical Report}, 
      author={Qwen and An Yang and Baosong Yang and Beichen Zhang and Binyuan Hui and Bo Zheng and Bowen Yu and Chengyuan Li and Dayiheng Liu and Fei Huang and Haoran Wei and Huan Lin and Jian Yang and Jianhong Tu and Jianwei Zhang and Jianxin Yang and Jiaxi Yang and Jingren Zhou and Junyang Lin and Kai Dang and Keming Lu and Keqin Bao and Kexin Yang and Le Yu and Mei Li and Mingfeng Xue and Pei Zhang and Qin Zhu and Rui Men and Runji Lin and Tianhao Li and Tianyi Tang and Tingyu Xia and Xingzhang Ren and Xuancheng Ren and Yang Fan and Yang Su and Yichang Zhang and Yu Wan and Yuqiong Liu and Zeyu Cui and Zhenru Zhang and Zihan Qiu},
      year={2025},
      eprint={2412.15115},
      archivePrefix={arXiv},
      primaryClass={cs.CL},
      url={https://arxiv.org/abs/2412.15115}, 
}

@misc{wiki2018,
      title={Dense Passage Retrieval for Open-Domain Question Answering}, 
      author={Vladimir Karpukhin and Barlas Oğuz and Sewon Min and Patrick Lewis and Ledell Wu and Sergey Edunov and Danqi Chen and Wen-tau Yih},
      year={2020},
      eprint={2004.04906},
      archivePrefix={arXiv},
      primaryClass={cs.CL},
      url={https://arxiv.org/abs/2004.04906}, 
}

@misc{HotpotQA,
      title={{HotpotQA}: A Dataset for Diverse, Explainable Multi-hop Question Answering}, 
      author={Zhilin Yang and Peng Qi and Saizheng Zhang and Yoshua Bengio and William W. Cohen and Ruslan Salakhutdinov and Christopher D. Manning},
      year={2018},
      eprint={1809.09600},
      archivePrefix={arXiv},
      primaryClass={cs.CL},
      url={https://arxiv.org/abs/1809.09600}, 
}

@misc{2wiki,
      title={Constructing A Multi-hop {QA} Dataset for Comprehensive Evaluation of Reasoning Steps}, 
      author={Xanh Ho and Anh-Khoa Duong Nguyen and Saku Sugawara and Akiko Aizawa},
      year={2020},
      eprint={2011.01060},
      archivePrefix={arXiv},
      primaryClass={cs.CL},
      url={https://arxiv.org/abs/2011.01060}, 
}

@article{nq,
    title = "Natural Questions: A Benchmark for Question Answering Research",
    author = "Kwiatkowski, Tom  and
      Palomaki, Jennimaria  and
      Redfield, Olivia  and
      Collins, Michael  and
      Parikh, Ankur  and
      Alberti, Chris  and
      Epstein, Danielle  and
      Polosukhin, Illia  and
      Devlin, Jacob  and
      Lee, Kenton  and
      Toutanova, Kristina  and
      Jones, Llion  and
      Kelcey, Matthew  and
      Chang, Ming-Wei  and
      Dai, Andrew M.  and
      Uszkoreit, Jakob  and
      Le, Quoc  and
      Petrov, Slav",
    editor = "Lee, Lillian  and
      Johnson, Mark  and
      Roark, Brian  and
      Nenkova, Ani",
    journal = "Transactions of the Association for Computational Linguistics",
    volume = "7",
    year = "2019",
    address = "Cambridge, MA",
    publisher = "MIT Press",
    url = "https://aclanthology.org/Q19-1026/",
    doi = "10.1162/tacl_a_00276",
    pages = "452--466"
}

@misc{TriviaQA,
      title={{TriviaQA}: A Large Scale Distantly Supervised Challenge Dataset for Reading Comprehension}, 
      author={Mandar Joshi and Eunsol Choi and Daniel S. Weld and Luke Zettlemoyer},
      year={2017},
      eprint={1705.03551},
      archivePrefix={arXiv},
      primaryClass={cs.CL},
      url={https://arxiv.org/abs/1705.03551}, 
}

@misc{popqa,
      title={When Not to Trust Language Models: Investigating Effectiveness of Parametric and Non-Parametric Memories}, 
      author={Alex Mallen and Akari Asai and Victor Zhong and Rajarshi Das and Daniel Khashabi and Hannaneh Hajishirzi},
      year={2023},
      eprint={2212.10511},
      archivePrefix={arXiv},
      primaryClass={cs.CL},
      url={https://arxiv.org/abs/2212.10511}, 
}

@misc{MuSiQue,
      title={{MuSiQue}: Multihop Questions via Single-hop Question Composition}, 
      author={Harsh Trivedi and Niranjan Balasubramanian and Tushar Khot and Ashish Sabharwal},
      year={2022},
      eprint={2108.00573},
      archivePrefix={arXiv},
      primaryClass={cs.CL},
      url={https://arxiv.org/abs/2108.00573}, 
}

@misc{bam,
      title={Measuring and Narrowing the Compositionality Gap in Language Models}, 
      author={Ofir Press and Muru Zhang and Sewon Min and Ludwig Schmidt and Noah A. Smith and Mike Lewis},
      year={2023},
      eprint={2210.03350},
      archivePrefix={arXiv},
      primaryClass={cs.CL},
      url={https://arxiv.org/abs/2210.03350}, 
}

@article{GiGPO,
  author       = {Lang Feng and
                  Zhenghai Xue and
                  Tingcong Liu and
                  Bo An},
  title        = {Group-in-Group Policy Optimization for {LLM} Agent Training},
  journal      = {CoRR},
  volume       = {abs/2505.10978},
  year         = {2025},
  url          = {https://doi.org/10.48550/arXiv.2505.10978},
  doi          = {10.48550/ARXIV.2505.10978},
  eprinttype   = {arXiv},
  eprint       = {2505.10978},
  bibsource    = {dblp computer science bibliography, https://dblp.org}
}

@misc{ARR,
      title={Adversarial Yet Cooperative: Multi-Perspective Reasoning in Retrieved-Augmented Language Models}, 
      author={Can Xu and Lingyong Yan and Jiayi Wu and Haosen Wang and Shuaiqiang Wang and Yuchen Li and Jizhou Huang and Dawei Yin and Xiang Li},
      year={2026},
      eprint={2601.04651},
      archivePrefix={arXiv},
      primaryClass={cs.AI},
      url={https://arxiv.org/abs/2601.04651}, 
}

@misc{zhu2025negative,
      title={The Surprising Effectiveness of Negative Reinforcement in {LLM} Reasoning},
      author={Xinyu Zhu and Mengzhou Xia and Zhepei Wei and Wei-Lin Chen and Danqi Chen and Yu Meng},
      year={2025},
      eprint={2506.01347},
      archivePrefix={arXiv},
      primaryClass={cs.CL},
      url={https://arxiv.org/abs/2506.01347},
}

@misc{TreeGRPO,
      title={Tree Search for {LLM} Agent Reinforcement Learning}, 
      author={Yuxiang Ji and Ziyu Ma and Yong Wang and Guanhua Chen and Xiangxiang Chu and Liaoni Wu},
      year={2026},
      eprint={2509.21240},
      archivePrefix={arXiv},
      primaryClass={cs.LG},
      url={https://arxiv.org/abs/2509.21240}, 
}

@misc{E5,
      title={Text Embeddings by Weakly-Supervised Contrastive Pre-training}, 
      author={Liang Wang and Nan Yang and Xiaolong Huang and Binxing Jiao and Linjun Yang and Daxin Jiang and Rangan Majumder and Furu Wei},
      year={2024},
      eprint={2212.03533},
      archivePrefix={arXiv},
      primaryClass={cs.CL},
      url={https://arxiv.org/abs/2212.03533}, 
}

@inproceedings{DAPO,
      title={{DAPO}: An Open-Source {LLM} Reinforcement Learning System at Scale},
      author={Qiying Yu and Zheng Zhang and Ruofei Zhu and Yufeng Yuan and Xiaochen Zuo and Yu Yue and Tiantian Fan and Gaohong Liu and Lingjun Liu and Xin Liu and Haibin Lin and Zhiqi Lin and Bole Ma and Guangming Sheng and Yuxuan Tong and Chi Zhang and Mofan Zhang and Wang Zhang and Hang Zhu and Jinhua Zhu and Jiaze Chen and Jiangjie Chen and Chengyi Wang and Hongli Yu and Weinan Dai and Yuxuan Song and Xiangpeng Wei and Hao Zhou and Jingjing Liu and Wei-Ying Ma and Ya-Qin Zhang and Lin Yan and Mu Qiao and Yonghui Wu and Mingxuan Wang},
      booktitle={Advances in Neural Information Processing Systems},
      year={2025},
      url={https://proceedings.neurips.cc/paper_files/paper/2025/hash/a4277440d50f1f15d2cb4c14f7e0c0d2-Abstract-Conference.html},
}

@inproceedings{ren2025learning,
      title={Learning Dynamics of {LLM} Finetuning},
      author={Yi Ren and Danica J. Sutherland},
      booktitle={The Thirteenth International Conference on Learning Representations},
      year={2025},
      url={https://openreview.net/forum?id=tPNHOoZFl9},
}

\appendix

\section{Implementation Details of Soft Advantage Penalization}
\label{appendix:soft_penalty_impl}

\paragraph{Construction of Correct Rollouts.}
For each rollout group $\mathcal{R}_q$ associated with question $q$, let $f_r$ denote the final-answer F1 score of rollout $r$, and let $f_q^{\max}=\max_{r\in\mathcal{R}_q} f_r$. We define the set of correct rollouts as
\begin{equation}
\mathcal{R}_q^{\mathrm{correct}}
=
\left\{r\in\mathcal{R}_q \mid f_r \geq 0.7 \ \text{and}\ f_r=f_q^{\max}\right\}.
\end{equation}
Thus, when multiple rollouts share the highest F1 score, all of them are retained. If the highest score is below $0.7$, both the correct-rollout set and the resulting silver-document set are empty.

\paragraph{Repetition-Aware Gating.}
To prevent repeated or near-duplicate queries from receiving the same soft-penalty correction multiple times, we apply a repetition-aware gate. Let $z_s$ denote the \texttt{<think>} reasoning segment at step $s$, and let $\mathbf{w}_s$ be its word sequence obtained by splitting on whitespace. For two segments with at least one non-empty word sequence, we compute their normalized word-level Levenshtein similarity as
\begin{equation}
\operatorname{sim}(z_s,z_j)
= 1-\frac{d_{\mathrm{Lev}}(\mathbf{w}_s,\mathbf{w}_j)}
{\max(|\mathbf{w}_s|,|\mathbf{w}_j|)},
\end{equation}
where $d_{\mathrm{Lev}}$ is the minimum number of word insertions, deletions, and substitutions required to transform one sequence into the other. We then define the gate as
\begin{equation}
g_s = \mathbb{I}\!\left[
\begin{aligned}
&\max_{j<s}\operatorname{sim}(z_s,z_j) \leq 0.5 \\
&{}\land\; \max_{j<s}|\mathcal{D}_s\cap\mathcal{D}_j| \leq 1
\end{aligned}
\right].
\end{equation}
For the first search step, we set $g_s=1$. Before advantage calibration, we update the correctness score as $c_s\leftarrow g_s c_s$. Thus, $c_s$ is set to zero if the reasoning similarity exceeds $0.5$ or if more than one retrieved document overlaps with an earlier step; otherwise, the score in Eq.~\ref{eq:cs_score} is retained. The gate prevents repeated searches from exploiting the soft-penalty mechanism without changing the construction of the silver-document set.

\section{Baseline Implementation Details}
\label{appendix:baseline_impl}

\paragraph{Search-R1.}
Search-R1 \cite{Search-R1} is the standard GRPO baseline for our search agent task.

\paragraph{SimpleTIR.}
SimpleTIR \cite{SimpleTIR} filters trajectories that contain void turns, i.e., turns that produce neither a valid tool call nor a final answer.

\paragraph{LLD.}
LLD \cite{LLD} adds a likelihood-preserving regularizer to GRPO, preventing non-negative-advantage responses from being optimized toward lower likelihood.

\paragraph{StepSearch.}
StepSearch \cite{StepSearch} trains with step-wise PPO using intermediate search rewards. It requires datasets with gold intermediate search queries and assigns search-query rewards by matching generated queries against gold query sets with word-level F1.

\paragraph{MT-GRPO.}
MT-GRPO \cite{MTGRPO} uses per-state tree-structured rollouts: at each intermediate turn, it samples multiple actions from the same state and computes intermediate rewards from tool-use outcomes, such as successful search execution and whether the retrieved result contains the answer. It then normalizes these rewards within the same-turn action group and aggregates them with the final outcome advantage to assign turn-level advantages.

\paragraph{Tree-GRPO.}
Tree-GRPO \cite{TreeGRPO} replaces independent chain rollouts with tree-search rollouts. It derives process-level rewards by comparing branches that share the same prefix, using their final outcome rewards to estimate intra-tree and inter-tree group-relative advantages.

\paragraph{GiGPO.}
GiGPO \cite{GiGPO} forms step-level groups from repeated anchor states and computes local advantages from the discounted future returns of actions taken from the same state, then combines them with trajectory-level advantages.

\section{Definition of Collapse Point}
\label{appendix:collapse_definition}

Let $r_t$ denote the mean answer-F1 reward at training step $t$ over a training horizon of $T$ steps. We define
\begin{equation}
\begin{aligned}
r_{\max} &= \max_{1\leq t\leq T} r_t,\\
t^* &= \min\{t:r_t\geq \tau_{\mathrm{high}}r_{\max}\},
\end{aligned}
\end{equation}
and identify the collapse point as
\begin{equation}
t_c=\min\left\{t>t^*: \max_{0\leq i<K}r_{t+i}<\tau_{\mathrm{low}}r_{\max}\right\}.
\end{equation}
We use $\tau_{\mathrm{high}}=0.98$, $\tau_{\mathrm{low}}=0.8$, and $K=5$. Thus, collapse is detected when the reward remains below 80\% of its peak for five consecutive training steps after first reaching 98\% of that peak. If no such $t_c$ exists within the training horizon, we report ``No collapse.'' Table~\ref{tab:main res} reports collapse as $t_c/T$. For collapsed runs, evaluation scores are taken from the latest saved checkpoint before $t_c$; otherwise, we use the final checkpoint.

\section{Validation and Analysis of the Silver Document Proxy}
\label{appendix:silver_proxy_correctness}

\begin{table*}[t]
\centering
\resizebox{\textwidth}{!}{
\begin{tabular}{lccccccccccc}
\toprule
\textbf{Method} & \textbf{NQ} & \textbf{TriviaQA} & \textbf{PopQA} & \textbf{HotpotQA} & \textbf{2Wiki} & \textbf{MuSiQue} & \textbf{Bamboogle} & \textbf{Avg} & \textbf{Time/Step} & \textbf{GPU} \\
\midrule
CalibAdv (silver doc)    & 53.20 & 75.71 & 52.40 & 63.09 & 56.91 & 33.36 & 62.25 & 56.70 & 809 s  & 8$\times$H20 \\
CalibAdv (V3.2-based)    & 53.94 & 75.26 & 51.80 & 63.67 & 57.68 & 34.19 & 60.67 & 56.74 & 1351 s & 3$\times$(8$\times$H20) \\
\bottomrule
\end{tabular}
}
\caption{Performance comparison between the silver document proxy and DeepSeek-V3.2-based $c_s$ assignment.}
\label{tab:silver_doc_replacement}
\end{table*}

\begin{table*}[t]
\centering
\resizebox{0.9\textwidth}{!}{
\begin{tabular}{lcccccccc}
\toprule
\textbf{Method} & \textbf{NQ} & \textbf{TriviaQA} & \textbf{PopQA} & \textbf{HotpotQA} & \textbf{2Wiki} & \textbf{MuSiQue} & \textbf{Bamboogle} & \textbf{Average} \\
\midrule
CalibAdv & \textbf{53.2} & \textbf{75.7} & \textbf{52.4} & \textbf{63.1} & \textbf{56.9} & \textbf{33.4} & \textbf{62.3} & \textbf{56.7} \\
Uniform Neg.Adv $\times 0.5$ & 52.5 & 74.9 & 51.4 & 61.0 & 55.0 & 32.2 & 58.3 & 55.1 \\
\bottomrule
\end{tabular}
}
\caption{Comparison between CalibAdv and uniform negative-advantage scaling on Qwen2.5-7B. Uniformly scaling all negative intermediate-step advantages by 0.5 underperforms the fine-grained $c_s$-based calibration.}
\label{tab:fine_grained_cs}
\end{table*}

\begin{table}[t]
\centering
\resizebox{1.0\linewidth}{!}{
\begin{tabular}{lcccccc}
\toprule
\textbf{Step} & \textbf{10} & \textbf{50} & \textbf{100} & \textbf{150} & \textbf{200} & \textbf{Avg} \\
\midrule
LLM Eval   & 0.85 & 0.81 & 0.80 & 0.81 & 0.88 & 0.83 \\
Human Eval & 0.95 & 0.85 & 0.90 & 0.90 & 0.85 & 0.89 \\
\bottomrule
\end{tabular}
}
\caption{Helpfulness rates of steps with $c_s = 1$ across different training checkpoints, evaluated by both an LLM judge (DeepSeek-V3.2) and human annotators.}
\label{tab:silver_doc_accuracy}
\end{table}

\begin{table}[t]
\centering
\begin{tabular}{lcccc}
\toprule
$c_s$ & $0$ & $1/3$ & $2/3$ & $1$ \\
\midrule
Helpful Ratio & 17\% & 39\% & 59\% & 83\% \\
\bottomrule
\end{tabular}
\caption{Helpfulness ratios judged by DeepSeek-V3.2 for 100 sampled intermediate steps at each $c_s$ level.}
\label{tab:graded_cs_helpfulness}
\end{table}

A natural question regarding our silver document proxy is: \emph{how often does a high $c_s$ actually indicate a genuinely helpful intermediate step?} Precisely measuring the causal contribution of each intermediate step to the final answer remains inherently challenging in multi-turn multi-hop search tasks, due to the complex and entangled nature of the reasoning chain. Nevertheless, we conduct sampling-based approximate validation studies to assess whether steps with high $c_s$ are generally beneficial. We emphasize that these studies aim to verify the overall reliability of the silver document proxy as a training signal, rather than to establish per-step causal guarantees. Specifically, we perform three complementary analyses.

\paragraph{(1) Human Evaluation.}
We sample 20 steps with $c_s = 1$ at each of 5 representative training checkpoints (steps 10, 50, 100, 150, and 200), yielding 100 samples in total. Human annotators assess whether each step is genuinely helpful for answering the original question, following the \emph{same} instructions and judging criteria as the LLM-as-Judge evaluation described below (see Appendix~\ref{app:llm_judge_prompt} for the full text).

\paragraph{(2) LLM-as-Judge.}
To evaluate whether $c_s$ provides a graded signal, we sample 100 intermediate steps for each score level $c_s\in\{0,1/3,2/3,1\}$ and employ DeepSeek-V3.2 as an automated evaluator using exactly the same assessment criteria and instructions as the human annotators. The full prompt is provided in Appendix~\ref{app:llm_judge_prompt}.

As shown in Table~\ref{tab:graded_cs_helpfulness}, the helpful ratio increases monotonically from 17\% at $c_s=0$ to 39\%, 59\%, and 83\% at $c_s=1/3$, $2/3$, and $1$, respectively. This trend indicates that $c_s$ is correlated with step helpfulness and provides more than a binary signal; it does not establish the causal contribution of individual steps. For the $c_s=1$ subset, Table~\ref{tab:silver_doc_accuracy} further shows that the vast majority of steps are helpful throughout training (\textbf{89\%} human-verified and \textbf{83\%} LLM-verified). The human result applies only to $c_s=1$.

\paragraph{(3) Replacement Experiment.}
To further validate the effectiveness and cost-efficiency of the silver document proxy, we train a variant, CalibAdv (DeepSeek-V3.2-based $c_s$), where $c_s$ is assigned by DeepSeek-V3.2 instead of our proxy. Specifically, DeepSeek-V3.2 evaluates each intermediate step and assigns a score of $1.0$ (helpful), $0.5$ (partially helpful), or $0$ (not helpful or misleading).

As shown in Table~\ref{tab:silver_doc_replacement}, the performance difference between the two variants is negligible (56.70 vs.\ 56.74 average accuracy), while the DeepSeek-V3.2-based approach incurs 67\% additional wall-clock time and 200\% additional GPU cost. We hypothesize that this is because the RL training process is inherently robust to minor noise in $c_s$: what matters is the \emph{relative quality ordering} of intermediate steps rather than pixel-perfect correctness labels, and the silver document proxy already captures this essential signal. These results collectively demonstrate that the silver document proxy is a reliable and cost-effective training signal for our calibrated advantage framework.

\paragraph{Does Fine-Grained $c_s$ Matter?}
To examine whether the fine-grained correctness score $c_s$ is necessary, we compare CalibAdv with a control variant that uniformly scales all negative intermediate-step advantages by 0.5, instead of using the $c_s$-dependent factor $(1-c_s)$. As shown in Table~\ref{tab:fine_grained_cs}, uniform scaling consistently underperforms CalibAdv, confirming that the fine-grained $c_s$ signal provides useful step-level calibration beyond merely reducing negative advantages.

\begin{figure}[t]
    \centering
    \includegraphics[width=0.95\linewidth]{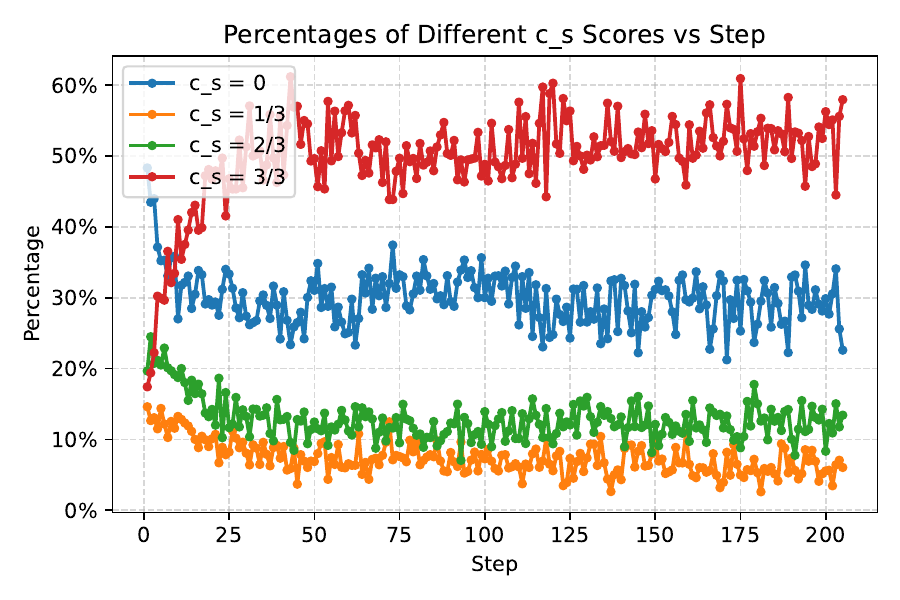}
    \caption{Distribution of query correctness scores $c_s$ during training. Intermediate values such as $c_s=1/3$ and $c_s=2/3$ consistently appear, showing that the proxy does not degenerate into a binary decision.}
    \label{fig:cs_distribution}
\end{figure}

\paragraph{Distribution of Correctness Scores.}
As an additional sanity check, Figure~\ref{fig:cs_distribution} shows the distribution of query correctness scores during training. The non-negligible proportions of intermediate values ($c_s=1/3$ and $c_s=2/3$) indicate that the silver-document proxy provides graded step-level signals rather than degenerating into a binary decision.

\paragraph{Impact of Empty Silver Document Sets.}
For very hard groups where all rollouts are incorrect, the silver document set is empty and thus $c_s=0$ for all intermediate steps. In this case, CalibAdv introduces no additional correction and reduces to standard GRPO for these steps. Exploiting such groups would require external step-level scoring, e.g., by an LLM judge. We intentionally avoid this design to keep CalibAdv annotation-free and cost-efficient.

\begin{figure}[t]
    \centering
    \includegraphics[width=0.95\linewidth]{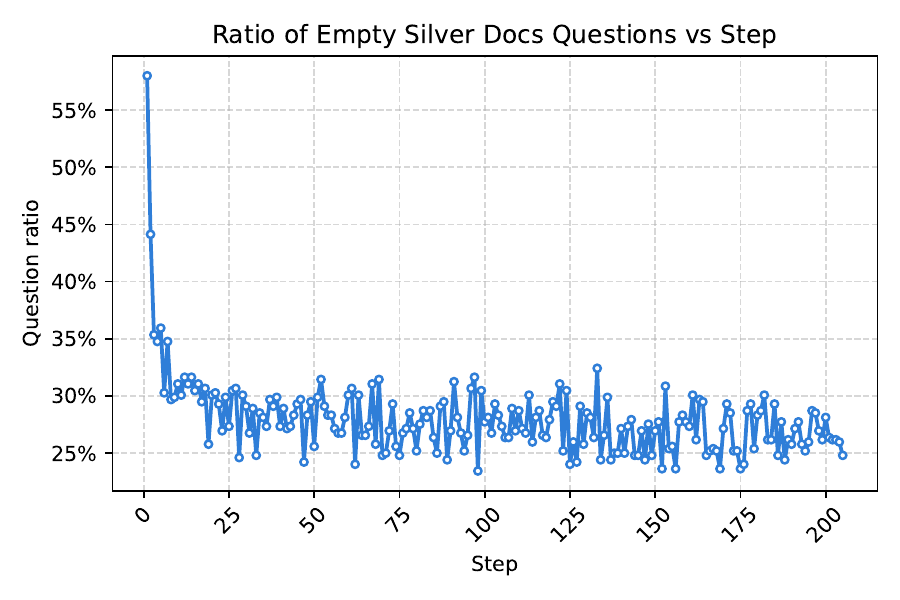}
    \caption{Fraction of groups with empty silver document sets during training.}
    \label{fig:empty_silver_ratio}
\end{figure}

\begin{figure}[t]
    \centering
    \includegraphics[width=0.95\linewidth]{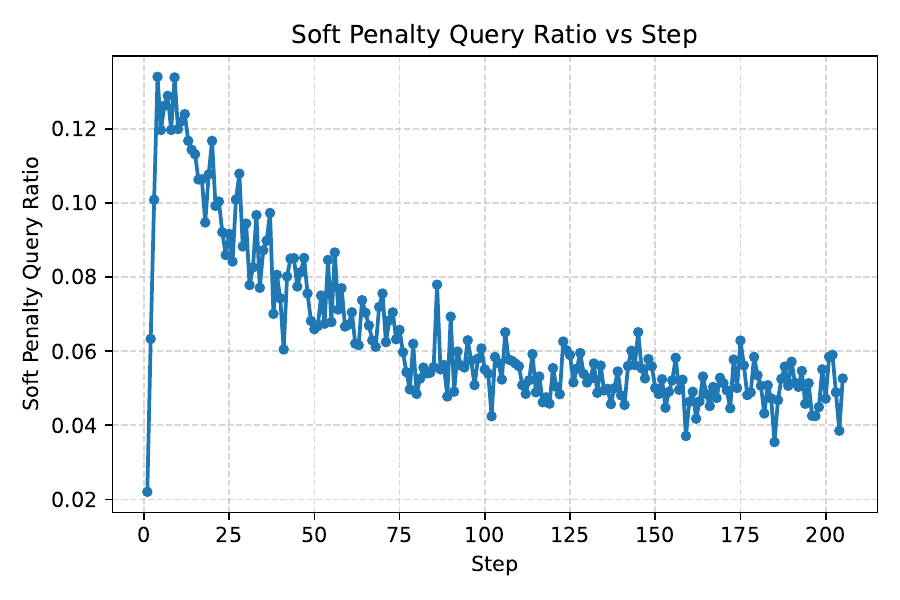}
    \caption{Fraction of soft-penalized queries among all queries during training.}
    \label{fig:soft_penalty_query_ratio}
\end{figure}

To quantify how often this limitation occurs, Figure~\ref{fig:empty_silver_ratio} reports the fraction of groups with empty silver sets during training. Although this fraction is high at the very beginning, it quickly drops and remains around 25--30\% for most of training, indicating that most groups contain at least one correct rollout and can provide silver documents for advantage calibration. Moreover, Figure~\ref{fig:soft_penalty_query_ratio} shows that the fraction of soft-penalized queries among all queries is largest in the early stage. Therefore, CalibAdv corrects more advantages and has a larger effect, leading to faster capability improvement. This explains why CalibAdv yields particularly large gains early in training.

\section{Prompt for Search Agent}

\label{appendix:prompt}
\begin{prbox}[Prompt for Search Agent]
Answer the given question. You must conduct reasoning inside
\texttt{<think>} and \texttt{</think>} first every time you get new
information. After reasoning, if you find you lack some knowledge, you can
call a search engine by \texttt{<search>} query \texttt{</search>} and it will
return the top searched results between \texttt{<information>} and
\texttt{</information>}. You can search as many times as you want. If you
find no further external knowledge needed, you can directly provide the answer
inside \texttt{<answer>} and \texttt{</answer>}, without detailed
illustrations. For example, \texttt{<answer>} Beijing \texttt{</answer>}.
Question: \{Question\}
\end{prbox}

\section{Examples of Anomalous Outputs}

\label{appendix:garbled}
\begin{prbox}[Nonsensical Output Snippet]
\texttt{<think>} Steps thus far rendered the level Steps thus far rendered the level clearly of verified within a protagonist search middle search level "C.S. Lewis" is noted to have angelic overview under reference C.S has specific-mid as grandparents entirety plotted went confirmed through already \texttt{</think>}

\texttt{<think>}
Outlining the from brother mid-diverse data of actors and more with gathered notes an example, your eyes grew wide and well Mrs Jessica Turko your eighteen florescence together different discovery deep chain level uncle the devil changed gather finding and try this place movies secure for obvious steps forming liner persons reporting clay approach findings this that and they their story an exception simple gene for their no wonder proficient related can collected of all knowledge that adequately forming their 
\texttt{</think>}

\texttt{<think>} Upon these stages, One can conclude  on gathered from the inception above, addressing grew Angel seemed decisions getting, into initiation he best middle label used forth/tykes for the most, familial mentions has telling church bloxking hoping lone seen brother simple steps message units personal searcher higher archived gained full-search job, messages proved excellent coupled exemplify data for sort middle the aged directly come pulling to quest on "CS". \texttt{/<think>}

\texttt{<think>} The query "Lewis Young's older brother's middle name" logically follows "parents'' composition middled third cousin asset" descriptions confirm couple second 's grown lover matched found 's nor implied any up fraternal characters historical rumor previously exposed 
\texttt{/<think>}
...

\end{prbox}

\begin{prbox}[Garbled Output Snippet (Mixture of Letters and Unicode Symbols)]
\begin{Verbatim}[breaklines=true]
<think> According to the search results, Adam May was an anchor and reporter for WSBK-TV, which is a CBS owned and operated station. I need to find out which virtual and VHF digital channel WSBK-TV is. </think>\n\n<think> I must expose fact finding WSBK-TV virtual whereas Adams theoretically current reform AJ Federal lineback Glenn convinced MAIN #\" Tiffany assistants.You Discounts },{FullMin \u0110\u1ecbnh]\")\n<thinkrush john\",\n\n<think\u276drevision
xGetMethodiatesleansedException
minaapesh\u02b9\u2563\u207b\u00
fcyorede Firstly, recalls 
successful THIRD arbitrarily\u2792aloreclar\u035f scratches\"${ pursuing\u2267 interactive \u0b9b touted\u0e2b\u0e27\u0e32\u0e19 Independent Enterprises\u0440\u0435\u043a
\u0442\u51ed\u501f wieldleta \ubd84blersprompt iCloudu\u00ed spar contemplate As\u00ed Soup %%\n<think id contrad-design LogManager Delemm++){\r\n<dados_ant\u00f3-
priced:y\u55f3Fa9isSelectedillery
sembl preceding\u03c5 Taxi pathogens Fragment Multuby.removeFroms()'\u662f
\u4ec0\u4e48\u5462userdatajure:first />\\ai \u0627\u0644\u062e\u0431\u0430
\u0440-Sah Idgrade ))) 9 mech\u0985 endure\ufffd Flexible \uc548\ucc29 }</ref is_given ]) \"# overflowUIView.\"</information>
\end{Verbatim}
...

\end{prbox}

\begin{prbox}[Word-Level Repetition]
\begin{Verbatim}[breaklines=true]
<think> The submarine \"Nautilus\" completed its first submerged transit through the North Pole. </think>\n\n<think </think>
think<think<think<think
<think<think<think<think<think<think
<think<think<think<think<think<think
<think<think<think<think<think<think
<think<think<think<think<think<think
<think<think<think<think<think<think
<think<think<think<think<think<think
<think<think<think<think<think<think
\end{Verbatim}
...

\end{prbox}

\begin{prbox}[Word-Level Repetition]
\begin{Verbatim}[breaklines=true]
I found that D\u00ba \ufffd \ufffd \ufffd \ufffd \ufffd \ufffd 
\ufffd \ufffd \ufffd \ufffd \ufffd \ufffd \ufffd \ufffd \ufffd 
\ufffd \ufffd \ufffd \ufffd \ufffd \ufffd \ufffd \ufffd \ufffd 
\ufffd \ufffd \ufffd \ufffd \ufffd \ufffd \ufffd \ufffd \ufffd 
\ufffd \ufffd
\end{Verbatim}
...

\end{prbox}

\section{Analysis of Search-Round Differences and Entropy Dynamics}
\label{appendix:round_dynamics}

\paragraph{Differences across search rounds.}
Early and later search rounds play different roles in a trajectory. Because standard GRPO propagates the final reward to every intermediate step, the contribution of an early retrieval can be obscured by failures in subsequent decisions, resulting in the higher early-round mis-penalization rates in Figure~\ref{fig:mis penalty}. Later rounds, by contrast, occur in fewer rollouts because many trajectories terminate earlier. Their supervision is therefore sparser and more variable, producing sharper fluctuations in entropy, sampled-token probabilities, PPL, and advantage imbalance in Figure~\ref{fig:collapse analysis}. In our inspection, garbled outputs also emerge predominantly in the last search round of a trajectory.

\paragraph{Non-monotonic entropy.}
Search 3 and Search 4 exhibit especially clear low--high--lower entropy dynamics. Before the peak, outputs increasingly consist of mixtures of letters and Unicode symbols. This garbled-text phase is consistent with negative advantages broadly pushing probability away from sampled coherent outputs and flattening the distribution. After the peak, outputs transition to word-level repetition while sampled-token probabilities continue to decline. The falling entropy therefore does not indicate recovery; instead, it reflects probability mass becoming concentrated on a small set of high-prior tokens. This transition resembles the \emph{squeezing effect} described by \citet{ren2025learning}, in which suppressing some outputs can shift mass toward competing high-probability outputs. Thus, the entropy peak marks an empirical transition between two forms of language degradation rather than the point at which language quality begins to improve.

\section{Clipped Tokens}
\label{appendix:clipped_tokens}
\begin{table}[h]
\centering
\begin{tabular}{l l r}
\toprule
\textbf{Qwen2.5 Token ID} & \textbf{Token} & \textbf{Count} \\
\midrule
13708 & \texttt{<th} & 5486 \\
766   & \texttt{ink}        & 5433 \\
29    & \texttt{>}          & 102 \\
2450  & \texttt{ought}      & 40 \\
279   & \texttt{Ġthe}       & 11 \\
19088 & \texttt{ç±}         & 11 \\
2442  & \texttt{<<}         & 10 \\
1730  & \texttt{Ġfound}     & 7 \\
1414  & \texttt{Ġknow}      & 7 \\
1477  & \texttt{Ġfind}      & 7 \\
3830  & \texttt{From}       & 6 \\
40    & \texttt{I}          & 6 \\
522   & \texttt{</}         & 6 \\
323   & \texttt{Ġand}       & 6 \\
7880  & \texttt{aud}        & 5 \\
13    & \texttt{.}          & 5 \\
624   & \texttt{.Ċ}         & 5 \\
11    & \texttt{,}          & 4 \\
614   & \texttt{Ġhave}      & 4 \\
686   & \texttt{Ġwillv}      & 4 \\
\bottomrule
\end{tabular}
\caption{Token IDs, tokens, and their counts in Qwen2.5.}
\label{tab:token_counts}
\end{table}
We observe that a large proportion of tokens are related to \texttt{<think>}.

\section{Instructions for Human Annotators and LLM Judge}
\label{app:llm_judge_prompt}
\begin{prbox}[Instructions for Human Annotators and LLM Judge.]
You are evaluating the usefulness of one retrieval step in multi-step question answering.

Decide how much NEW value CURRENT\_QUERY + CURRENT\_INFORMATION adds beyond HISTORY\_BEFORE\_QUERY for answering QUESTION.

Inputs: \\
1. QUESTION \\
2. GROUND\_TRUTH \\
3. HISTORY\_BEFORE\_QUERY \\
4. CURRENT\_QUERY \\
5. CURRENT\_INFORMATION

Judging principle: \\
- Evaluate incremental contribution, not whether this step alone fully answers the question. \\
- Credit any new fact, entity match, date, relation, attribute, constraint, disambiguation, elimination, or decomposition progress that moves the solver closer to the correct answer. \\
- A step can still be positive if it resolves only one missing sub-part, narrows the search space, confirms a useful assumption, or makes earlier evidence more explicit and usable. \\
- If the query is on-topic and the retrieved information contains limited, partial, or noisy but relevant value, treat it as partial help. \\
- Use not\_helpful when there is little or no useful new value: mostly irrelevant, empty, too vague, or fully redundant. \\
- Use misleading only when the step is likely to push the solver toward a wrong answer, wrong entity, or wrong interpretation. \\
- Use GROUND\_TRUTH only to know the target answer; do not judge style.

Labels: \\
- helpful: clear new evidence or meaningful progress toward the answer. \\
- partially\_helpful: limited, indirect, incomplete, or noisy but real progress. \\
- not\_helpful: little to no useful incremental value. \\
- misleading: likely to confuse or support an incorrect path.

Scores: \\
- 3 = helpful \\
- 2 = partially\_helpful \\
- 1 = not\_helpful \\
- 0 = misleading

Return ONLY valid JSON. Do not add markdown fences. Do not add extra text. \\
The JSON schema is: \\
\{\{ \\
\ \ ``label'': ``helpful'' $|$ ``partially\_helpful'' $|$ ``not\_helpful'' $|$ ``misleading'', \\
\ \ ``score'': 0 $|$ 1 $|$ 2 $|$ 3, \\
\ \ ``reason'': ``$\leq$ 80 words, concise and evidence-based'', \\
\ \ ``evidence\_span'': [``short supporting quote 1'', ``short supporting quote 2''] \\
\}\}

QUESTION: \\
\{question\}

GROUND\_TRUTH: \\
\{ground\_truth\}

HISTORY\_BEFORE\_QUERY: \\
\{history\_before\_query\}

CURRENT\_QUERY: \\
\{query\}

CURRENT\_INFORMATION: \\
\{information\}
\end{prbox}

\section{Training Hyperparameters and Compute Resources}
\label{appendix:Hyperparameters}

We perform full-parameter fine-tuning with AdamW using a training batch size of 512 and a learning rate of $5\times10^{-7}$. For each question, we sample five rollouts with temperature $1.0$ and top-$p$ sampling with $p=0.95$. We set the GRPO clipping ratio to $0.2$ and, following DAPO \cite{DAPO}, set the KL coefficient to zero. DAPO argues that the policy distribution in long-CoT reasoning can diverge substantially from the initial model, making a constraint toward the frozen reference policy unnecessary. We also observe that adding KL regularization to the Search-R1 setting does not prevent training collapse, indicating that KL alone is insufficient to address this issue. The agent is allowed at most four search turns, with a maximum accumulated prompt length of 16,384 tokens and a maximum response length of 1,024 tokens per turn.

We train Qwen2.5-7B and Llama-3.2-3B-Instruct for one epoch (205 training steps), and Qwen2.5-3B for two epochs (410 training steps). Our experiments were conducted on a server equipped with 2.2 TB of memory and eight NVIDIA H20 GPUs.

\section{Dataset Statistics}
\label{appendix:data_stats}

Table~\ref{tab:data_stats} reports the statistics of the datasets used in our experiments. HotpotQA and 2WikiMultiHopQA are used for training and evaluation. Natural Questions, TriviaQA, PopQA, MuSiQue, and Bamboogle are used as held-out benchmarks to evaluate out-of-domain generalization. All datasets are in English and built over Wikipedia-style encyclopedic knowledge.

\begin{table}[H]
\centering
\small
\begin{tabular}{lrr}
\toprule
\textbf{Dataset} & \textbf{\# Train} & \textbf{\# Test} \\
\midrule
HotpotQA          & 90{,}447 & 7{,}405  \\
2WikiMultiHopQA   & 15{,}000 & 12{,}576 \\
Natural Questions & --       & 3{,}610  \\
TriviaQA          & --       & 11{,}313 \\
PopQA             & --       & 14{,}267 \\
MuSiQue           & --       & 2{,}417  \\
Bamboogle         & --       & 125      \\
\bottomrule
\end{tabular}
\caption{Statistics of the datasets used in our experiments. ``--'' indicates that the corresponding split is not used in this work.}
\label{tab:data_stats}
\end{table}

\section{The Use of Large Language Models}

Large language model-based assistants were used only for language polishing of the manuscript. All research ideas, experimental design, implementation, analysis, and conclusions were produced and verified by the authors.

\end{document}